\def\year{2022}\relax
\documentclass[letterpaper]{article} % DO NOT CHANGE THIS
\usepackage{aaai22}  % DO NOT CHANGE THIS
\usepackage{times}  % DO NOT CHANGE THIS
\usepackage{helvet}  % DO NOT CHANGE THIS
\usepackage{courier}  % DO NOT CHANGE THIS
\usepackage[hyphens]{url}  % DO NOT CHANGE THIS
\usepackage{graphicx} % DO NOT CHANGE THIS
\usepackage{natbib}  % DO NOT CHANGE THIS AND DO NOT ADD ANY OPTIONS TO IT
\usepackage{caption} % DO NOT CHANGE THIS AND DO NOT ADD ANY OPTIONS TO IT
\DeclareCaptionStyle{ruled}{labelfont=normalfont,labelsep=colon,strut=off} % DO NOT CHANGE THIS
\usepackage{booktabs}
\usepackage{amsmath}
\usepackage{amssymb}
\usepackage{amsthm}
\usepackage{physics}
\usepackage{numprint}
\usepackage{newfloat}

\copyrighttext{Presented at the AI-HRI Symposium at AAAI Fall Symposium Series (FSS) 2022}

\title{Observed Adversaries in Deep Reinforcement Learning}
\author{Eugene Lim and Harold Soh}
\affiliations{
    National University of Singapore\\
    13 Computing Drive\\
    Singapore 117417\\
    \{elimwj,hsoh\}@comp.nus.edu.sg
}

\begin{document}

\maketitle

\begin{abstract}
In this work, we point out the problem of observed adversaries for deep policies. Specifically, recent work has shown that deep reinforcement learning is susceptible to adversarial attacks where an observed adversary acts under  environmental constraints to invoke natural but adversarial observations. This setting is particularly relevant for HRI since HRI-related robots are expected to perform their tasks around and with other agents. In this work, we demonstrate that this effect persists even with low-dimensional observations. We further show that these adversarial attacks transfer across victims, which potentially allows malicious attackers to train an adversary without access to the target victim. 
\end{abstract}

\section{Introduction}

Recent years have seen a significant gain in robot capabilities, driven in-part by progress in artificial intelligence and machine learning. In particular, deep learning has emerged as a dominant methodology for crafting data-driven components in robot systems~\cite{punjani15,sergey16}. However, the \emph{robustness} of such methods have recently come under scrutiny. Specifically, concerns have been raised about the susceptibility of deep methods to adversarial attacks~\cite{szegedy13}. For example, recent work has shown that small optimized pixel perturbations can drastically change the predictions of computer vision models~\cite{szegedy13,goodfellow14}.  

In this work, we focus on deep reinforcement learning (DRL)~\cite{mnih15,schulman15,schulman17}, which has been used to obtain policies for various robot tasks, including those involving human-robot interaction (HRI)~\cite{modares16,khamassi18,xie21}. Early works~\cite{huang17} showed that adversarially modified inputs (similar to those used against computer vision models) can be detrimental to agent behavior. Recently, \citet{gleave20} demonstrated that artificial agents are vulnerable under a more realistic threat model: natural observations that occur as a result of an adversary's behavior under environmental constraints. These \textit{observed adversaries} are not able to arbitrarily modify the victim's inputs, yet are able to significantly affect the victim's behavior.

Here, we build upon \citet{gleave20} and show that the observed adversary attacks are potentially even more insidious. While it is natural to suspect that this vulnerability mainly stems from the faulty perception of high-dimensional observations, our experiments show that deep policies remain susceptible in \emph{low}-dimensional settings where the environmental state is \emph{fully-observed}. In other words, deep policies are not robust to observed adversaries even in arguably simple settings. We further show that an observed adversary can successfully attack previously unseen victims, which has broader downstream implications. %allows an malicious  to train an adversary even in the absence of our target victim.

In the following, we will first detail experiments designed to investigate observed adversary attacks. We focus on Proximal Policy Optimization (PPO) \cite{schulman17}, a popular model-free RL method that has been widely used, including for  HRI~\cite{xie21}. We then present our results related to the severity and transferrability of attacks. Finally, we discuss the implications on our findings on HRI and potential future work that is needed to address the robustness of deep RL and advance the development of trustworthy robots.

\section{Background \& Related Work}

There is a rich literature on adversarial attacks on machine learning algorithms. Famous examples include adversarial attacks on deep computer vision models~\cite{szegedy13,goodfellow14}. Typically, these attacks involve solving an optimization problem to find the smallest perturbation on image pixels that is required to raise the classification loss. White-box attacks such as Fast Gradient Sign Method (FGSM) and Projected Gradient Descent (PGD) approximate the solution to this optimization problem. This approach has been extended to black-box settings mainly by exploiting the transferability of adversarial examples \cite{papernot16,papernot16-2}.

Recently, \citet{huang17} showed that gradient-based attacks are also effective in RL settings. However, these attacks assume a powerful adversary who is able to directly modify the victim/robot's observations. \citet{gleave20} worked under a more realistic setting where the adversaries are just agents acting in a multi-agent environment alongside the victims. In their work, they train an adversary to act under the constraints of the environment (i.e., a zero-sum simulated robotics game) to invoke natural but adversarial observations. These observations, when seen by the pre-trained victims, can damage the victim's ability to perform its task. In this short paper, we contribute to this line of work and investigate the susceptibility of deep policies in low-dimensional settings. Specifically, we study the situation where an RL agent can perceive the world perfectly and ask: \emph{is this agent still susceptible to these observed adversary attacks?}

\begin{figure}[t]
\begin{center}
\includegraphics[width=0.9\columnwidth]{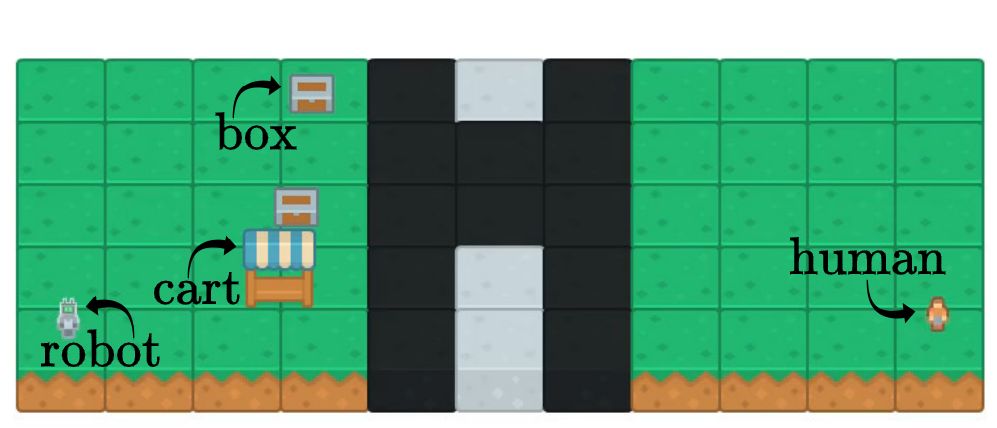}
\end{center}
\caption{Design of \texttt{twosides} level in Blockland environment. The robot aims to transfer all the boxes to the cart and the human is the observed adversary.}
\label{fig:blockland}
\end{figure}

\section{Experimental Setup}

% \hsnote{start by saying what the goal of your experiment is. Explain the high level idea.}

We designed our experiments to study the effect of observed adversaries when the state is low-dimensional and fully-observed. We are particularly interested in the severity and transferability of attacks across robot policies. At a high-level, our setup consists of training an adversary (simulated human agent) against a trained robot, and measuring the attacker's effect on the robot's accumulated reward. The environment and training code are available on \url{https://github.com/clear-nus/observersary}.

\begin{figure}[t]
\centering
\includegraphics[width=0.7\columnwidth]{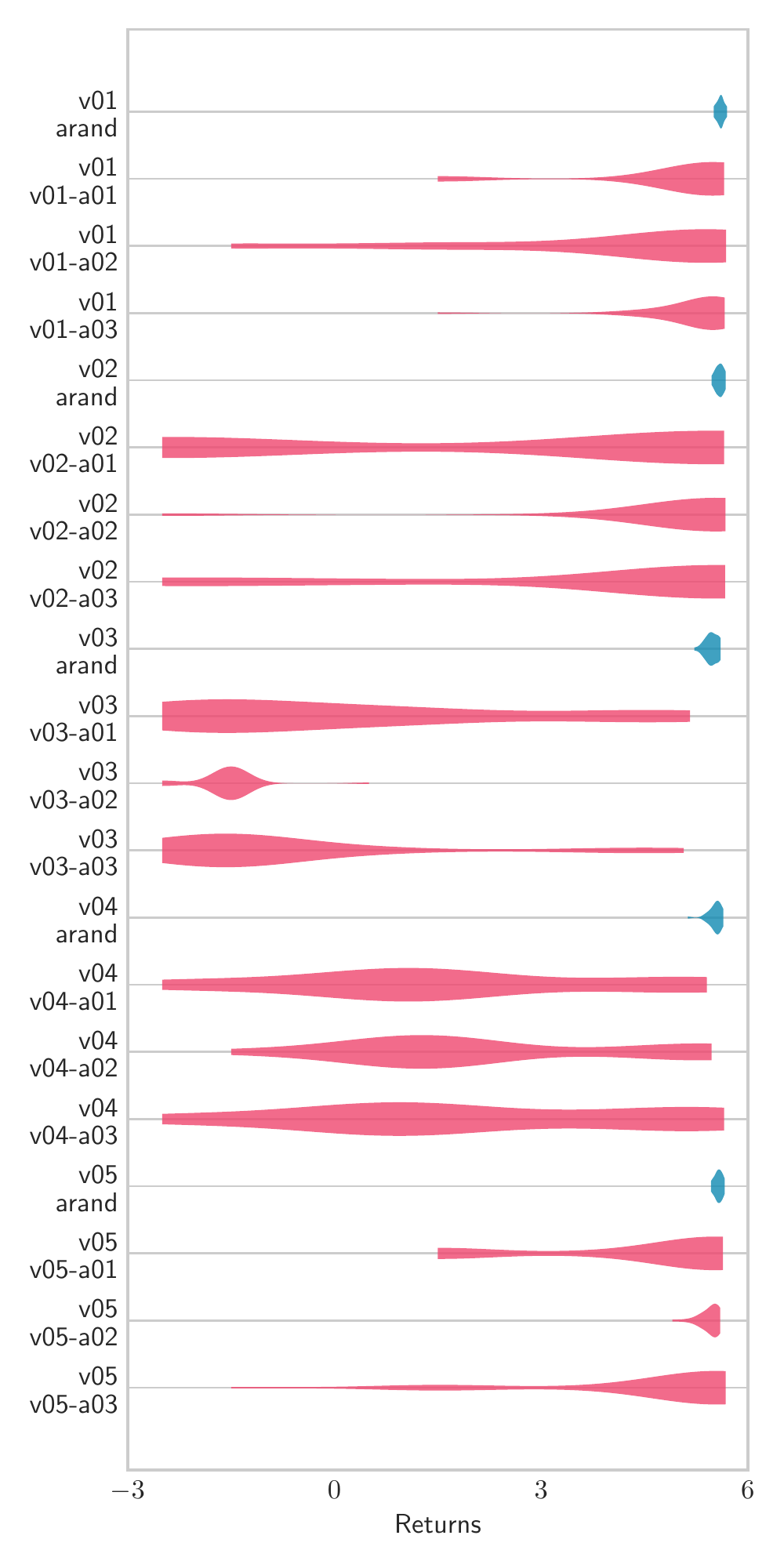}
\caption{Violin plot of the evaluation. Each row represents the distribution of returns (across 30 runs) attained by victim \texttt{v0x} against \texttt{arand} or its direct adversary \texttt{v0x-a0y}. Violins that are colored blue (resp. red) indicates that the victim is paired with \texttt{arand} (resp. an adversary).}
\label{fig:violin}
\end{figure}

\paragraph{Environment.} We conduct our experiments in \textit{Blockland}, an environment that that supports customizable levels. Figure \ref{fig:blockland} shows the design of the \texttt{twosides} level. At each time step, the human and robot can choose to move in one of four directions, do nothing, or interact with a nearby entity. An agent can pick up a box by interacting with it and place it down on the cart by interacting with the cart while holding the box. An agent can hold only one item at a time. Both agents can observe:
\begin{enumerate}
    \item the $xy$-positions of both agents,
    \item the $xy$-positions of both boxes,
    \item the $xy$-positions of the cart, and
    \item the number of boxes  that each agent is holding (0 or 1).
\end{enumerate}
Note that the positions are real-valued and continuous.

\paragraph{Tasks and Adversary.} The robot's goal is to transfer both boxes to the cart. It receives +1 reward for picking up a box and +2 reward for placing the box down on the cart. The maximum achievable episodic return is 6. To incentivize the speed of completion, the robot receives a -0.005 penalty per time step. The episode terminates when the robot has transferred both boxes to the cart or after 500 time-steps.

The human acts as the observed adversary in our experiment. To ensure that the human only influences the robot through observations, both agents are not allowed to cross the road.

\paragraph{Robot Training.} We train five robot with different random seeds to obtain ``victim'' policies \texttt{v01} to \texttt{v05}. All robots are trained for \numprint{800000} steps with a human who takes random actions with equal probability. A human that exhibits such random walk behavior is labelled \texttt{arand}. To ensure the experiments are reproducible, we trained all policies using \texttt{stable-baselines3} implementation of Proximal Policy Optimization (PPO). We choose to focus on PPO due to its popularity. Both actor and critic network are defined by the default \texttt{MlpPolicy} model that uses two 64-node hidden layers (please see the Appendix \ref{sec:hyperparameters} for the full list of hyperparameters used).

%\hsnote{why did you choose PPO? Why not other methods?}

\paragraph{Adversary Training.} For each victim \texttt{v0x}, we train 3 human adversaries to obtain adversarial policies \texttt{v0x-a01} to \texttt{v0x-a03}. An adversary is trained for \numprint{800000} steps and their objective is to maximize the negative reward of the robot. These are trained using \texttt{stable-baselines3} implementation of PPO with \texttt{MlpPolicy} model.

\begin{figure}[t]
\centering
\includegraphics[width=0.8\columnwidth]{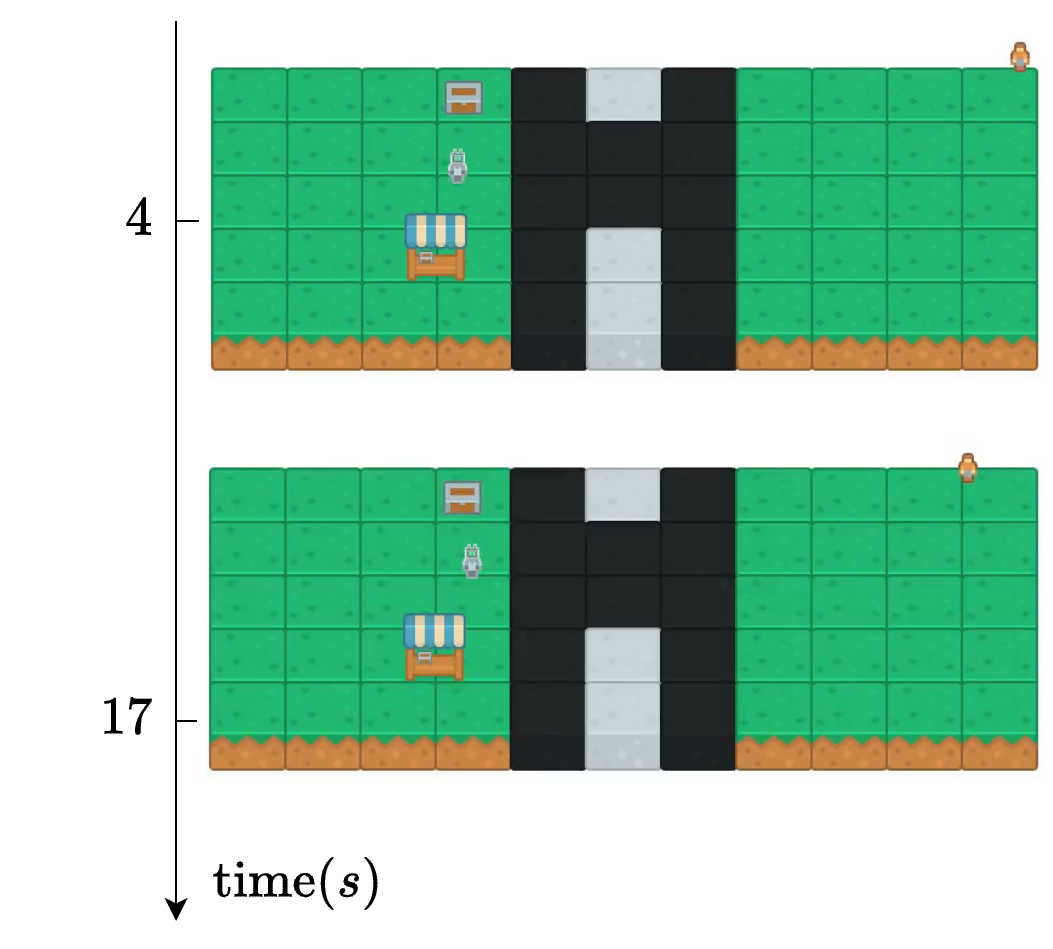}
\caption{Example of a typical failure by the victim. In this particular episode, the human adversary learns that loitering around the corner freezes the robot victim. The robot freezes while moving towards the second box, and remains frozen throughout the entire episode.}
\label{fig:failure}
\end{figure}

\section{Results}
In this section, we discuss our key findings. Our primary metric will be the robot's return (accumulated reward) at a test trial where an  adversary policy is paired with a specific robot and both are executed in the environment. 

\paragraph{Are agents susceptible to observed adversary attacks in settings with low-dimensional observations?}

%To answer this question, we train 5 robot victims with different seeds to obtain victim policies \texttt{v01} to \texttt{v05}. 
To answer this question, we evaluated each robot against \texttt{arand} and its three adversaries, each for 30 runs. Figure \ref{fig:violin} summarizes our results and shows that all robots are able to attain close to the maximum allowable returns in all 30 runs against \texttt{arand}. However, all three robots suffer a significant performance degradation against at least one of their direct adversaries, with their mean returns decreased by up to 126\%.

Qualitatively, we observe that performance degradation primarily occurs because the attacked robots stop moving if the human adversaries loiter around a small area near the edge of the map. Figure \ref{fig:failure} illustrates an example of a typical failed run by the victim. 

\begin{figure}[t]
\centering
\includegraphics[width=0.8\columnwidth]{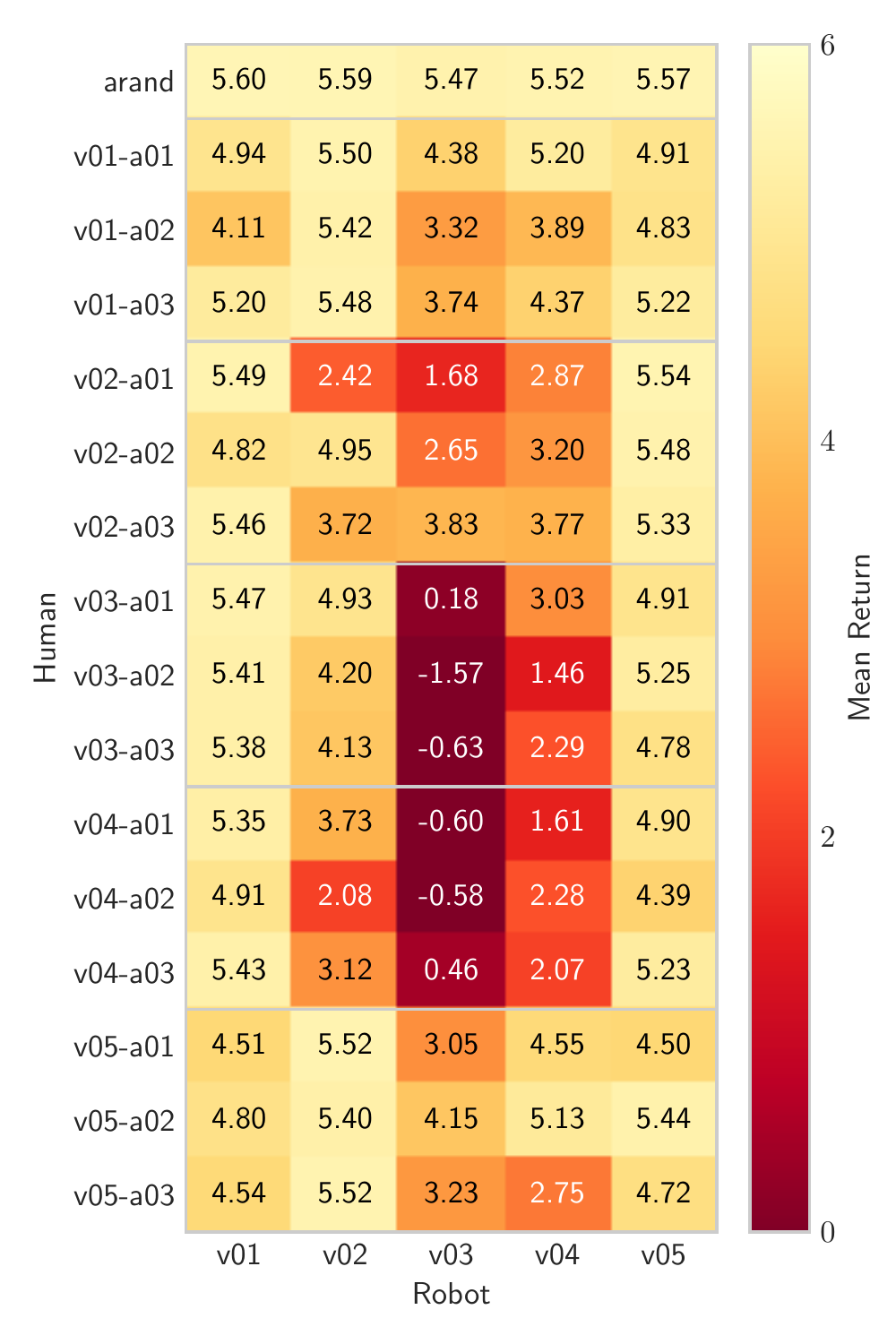}
\caption{Mean reward of the victim against \texttt{arand} and across the transfer adversaries.}
\label{fig:transfer}
\end{figure}

\paragraph{Does an adversary only work with the specific agent it is trained with or is it effective across victims?}

An attacker often does not have unrestricted access to the victim's policy, which renders the above attack infeasible in practice. In supervised learning settings, it has been shown that adversarial examples can be transferred across models~\cite{ian15}. Here, we investigate if this property also holds in reinforcement learning with observed adversaries.

We compare the performance of each robot victim \texttt{v0x} against ``transfer adversaries'', i.e., human agents \texttt{v0z-a0y} that are trained with a \emph{different} victim \texttt{v0z}$\neq$\texttt{v0x}. Figure \ref{fig:transfer} shows the mean returns of the robot victim across 30 runs against \texttt{arand} and various adversaries. We observe that an transfer adversary can have a negative effect on a robot. For example, \texttt{v03} returns fell by 6.05 against \texttt{v04-a02}, who was trained with robot \texttt{v04}. The effect is different across the robot policies, which suggests that the victim policies are indeed different. Surprisingly, the transfer adversary is sometimes able to inflict more damage to a victim than its direct training counterpart, as seen in \texttt{v02} against \texttt{v04-a02}. 

\begin{figure}[t]
\centering
\includegraphics[width=0.7\columnwidth]{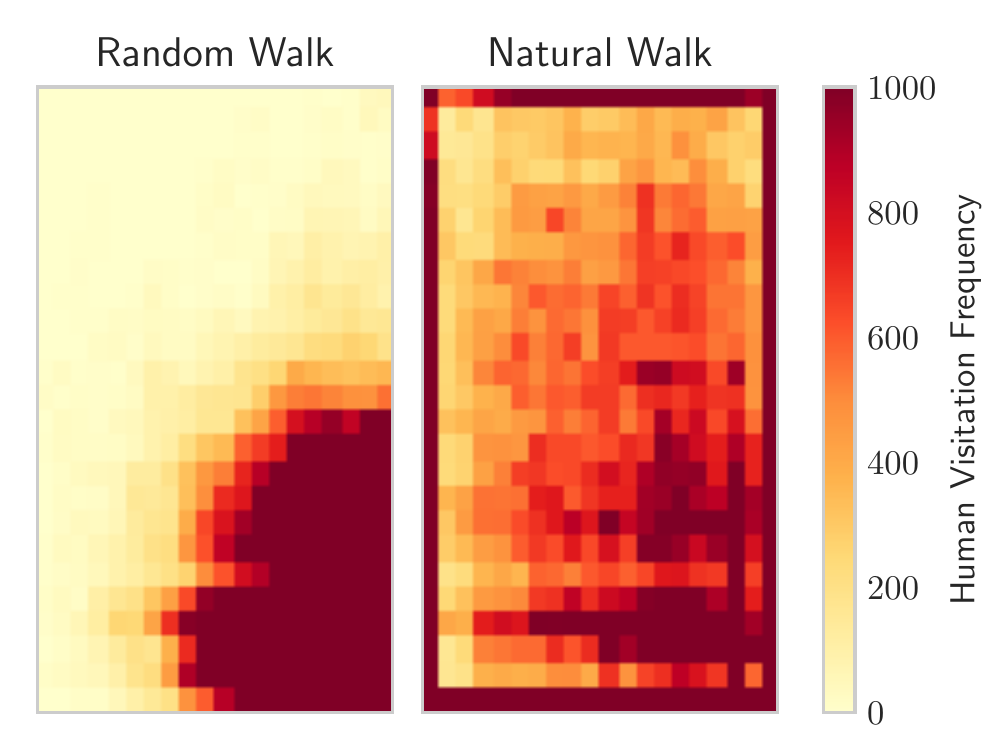}
\caption{Heat map of \texttt{arand} visitation frequency using different randomization.}
\label{fig:heatmap-twosides}
\bigskip
\includegraphics[width=0.7\columnwidth]{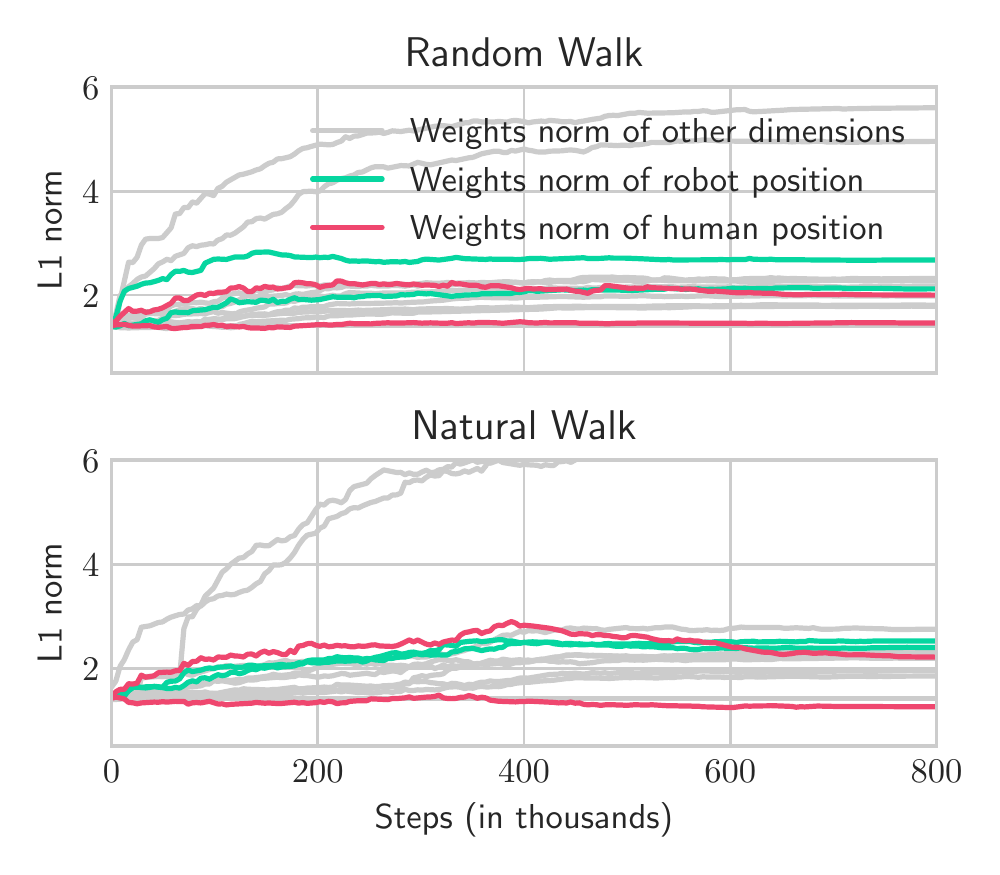}
\caption{L1 norms of the weights associated to each input dimensions.}
\label{fig:norms}
\end{figure}

\begin{figure}[t]
\centering
\includegraphics[width=0.8\columnwidth]{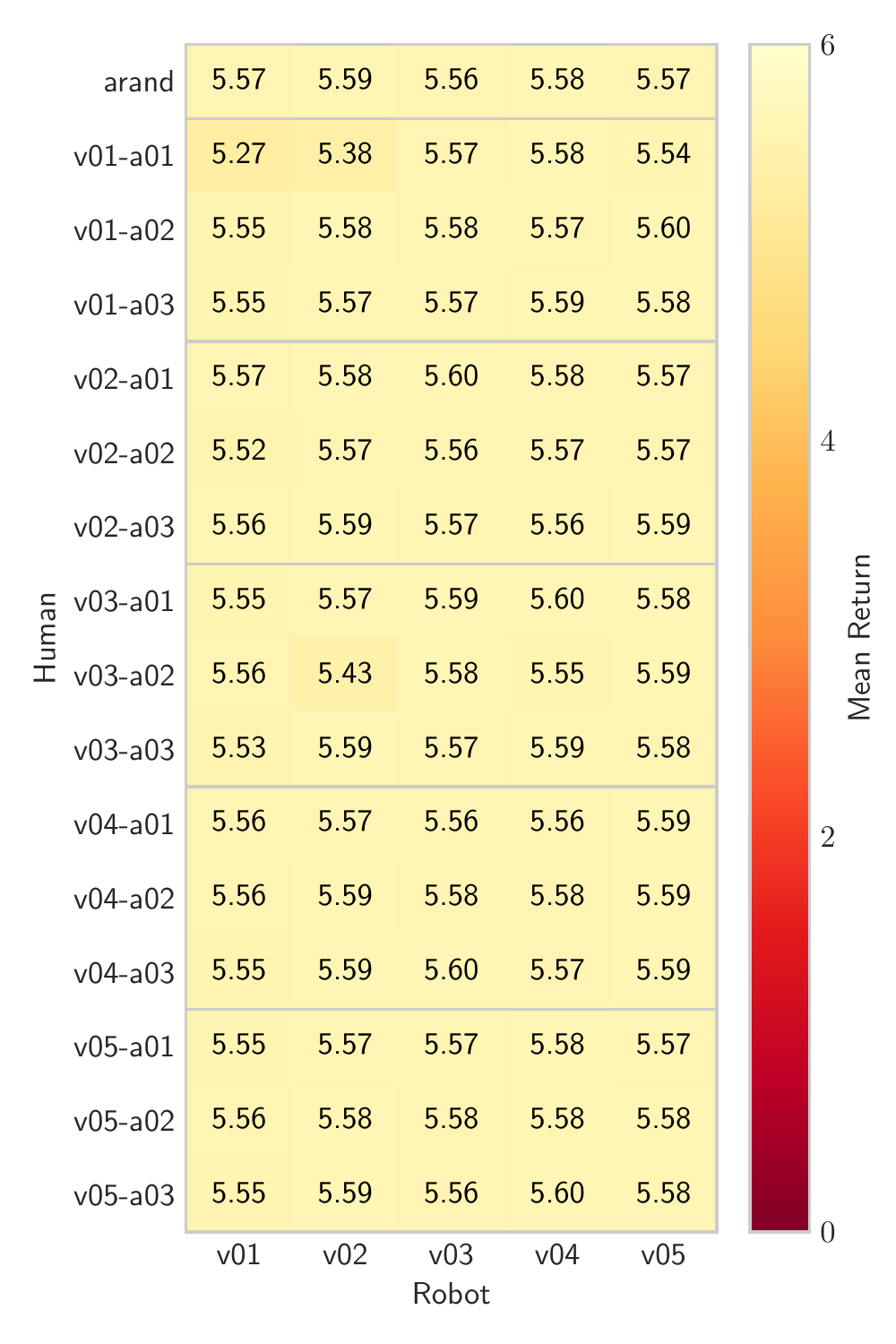}
\caption{Mean reward of the victim when trained with natural walk human agents.}
\label{fig:natural}
\end{figure}

\section{Discussion}

Our results above show that PPO-trained agents can be susceptible to observed adversaries even in relatively simple fully-observed environments. We are currently running experiments using Gym's Classic Control and Box2D environments, and preliminary results suggest that this attack persists across multiple environments. As such, even if a deep perception model is able to perfectly predict the state, the policy may still remain vulnerable. More significantly, observed adversaries are not only able to degrade training robots, but also unseen robot agents. This implies that malicious attackers may not require access to the victim policy to learn an adversarial policy. 

A natural question is why the robot policies are susceptible in this setting. Given that the human's position is completely irrelevant to task completion, the policy should learn to ignore those observation dimensions. An ``easy'' way to achieve this is to zero the input weights associated with irrelevant dimensions\footnote{It is possible to ignore irrelevant inputs without zeroing out the input weights, but this can be challenging since the upper layers would need to ``cancel out'' the irrelevant input's contributions.}. 
However, we find that this is not the case; Figure \ref{fig:norms} shows that the norms associated with the input layer weights of a robot policy as it is trained. Note that the norm of the weights associated with the human positions did not converge to zero. In other words, the policy has \emph{not} learnt to disregard the human observations at the input level.

The observed adversary can then exploit the victim by providing out-of-distribution observations. This raises a question of whether robot behavior can be made more robust if the training was performed with a different human agent that better explored the space. We ran an experiment to test this notion by using a ``natural walk'' policy; the human walks in a random direction for a random duration between 5 to 15 steps, repeatedly throughout an episode. Figure \ref{fig:heatmap-twosides} shows the heat map of the visitation frequency of both randomization schemes. Clearly, the natural walk covers the  accessible space more than the random walk.

Figure \ref{fig:natural} shows the performance of the victims against its adversaries when the victims are trained against natural walkers. The adversarial effect is no longer apparent. Interestingly, we see in Fig. \ref{fig:norms} that the policy under natural walk doesn't zero out the input weights associated with the human position, yet remains robust. It remains an open question why this is the case, but we hypothesize that the robot has learnt to respond correctly given the broad coverage of the observation space. Rigorous experimentation to validate this hypothesis remains future work. In addition, it is unclear how to construct effective training policies in general; it may be infeasible to construct policies that provide extensive coverage of the observation space in more complex settings.

To the best of our knowledge, observed adversary attacks remain relatively unexplored in the literature. As such, the precise mechanisms that permit this attack remains unknown. 
To effectively defend against such attacks, it is necessary to identify the root cause of the problem: is it due to a fundamental limitation of neural network policies or a quirk of the PPO algorithm? Experiments across various policies and algorithms are needed to address this question. 

Observed adversaries are particularly problematic for HRI, e.g., for social robots or collaborative robots that are expected to perform in environments with other (human) agents. As shown in this work, the attacker need not physically interact with a robot to cause significant performance degradation. Successful attacks on robots can drastically undermine trust and limit adoption. How to properly ensure that deep RL policies are robust before deployment remains an open question.

\appendix

\section{Training Hyperparameters}
\label{sec:hyperparameters}

The following set of hyperparameters are used when training the victims and adversaries.

\vspace{1em}
\begin{table}[h]
\centering
\begin{tabular}{rl}
\toprule
\textbf{Hyperparameter} & \textbf{Value} \\
\midrule
Number of environments & 8 \\
Steps per environment & 512 \\
Total time-steps & 800000 \\
Policy model & MlpPolicy \\
Learning rate & 0.001 \\
Batch size & 64 \\
Number of epochs & 10 \\
Discount factor $\gamma$ & 0.99 \\
GAE $\lambda$ & 0.95 \\
Clip range & 0.2 \\
Entropy coefficient & 0.01 \\
Value coefficient & 0.5 \\
Max grad norm & 0.5 \\
\bottomrule
\end{tabular}
\end{table}

\section*{Acknowledgments}

This research is supported by the National Research Foundation (NRF), Singapore and DSO National Laboratories under the AI Singapore Program (Award Number: AISG2-RP-2020-017). This research/project is supported by the National Research Foundation, Singapore under its AI Singapore Programme (AISG Award No: AISG-PhD/ 2021-08-011).

\bibliography{aaai22}

\end{document}